\documentclass{article}

\usepackage{PRIMEarxiv}

\usepackage[utf8]{inputenc} 
\usepackage[T1]{fontenc}    
\usepackage{hyperref}       
\usepackage{url}            
\usepackage{booktabs}       
\usepackage{amsfonts}       
\usepackage{nicefrac}       
\usepackage{microtype}      
\usepackage{lipsum}
\usepackage{fancyhdr}       
\usepackage{graphicx}       
\graphicspath{{media/}}     

\title{Developmental PreTraining (DPT) for Image Classification Networks}

\author{
  Niranjan Rajesh, Debayan Gupta \\
  Ashoka University \\
  Sonipat, India\\
  \texttt{\{niranjan.rajesh\_asp24, debayan.gupta\}@ashoka.edu.in} \\
}

\begin{document}
\maketitle

\begin{abstract}
In the backdrop of increasing data requirements of Deep Neural Networks for object recognition that is growing more untenable by the day, we present \textsc{Developmental PreTraining} (DPT) as a possible solution. DPT is designed as a curriculum-based pre-training approach designed to rival traditional pre-training techniques that are data-hungry. These training approaches also introduce unnecessary features that could be misleading when the network is employed in a downstream classification task where the data is sufficiently different from the pre-training data and is scarce. We design the curriculum for DPT by drawing inspiration from human infant visual development. DPT employs a phased approach where carefully-selected primitive and universal features like edges and shapes are taught to the network participating in our pre-training regime. A model that underwent the DPT regime is tested against models with randomised weights to evaluate the viability of DPT.
\end{abstract}


\section{Introduction}
The advent of Deep Learning (DL) has massively aided the Artificial Intelligence community, especially in the realm of object recognition. One of the critical reasons for the success of DL has been the availability of massive image datasets \cite{Deng2009ImageNet} and the computational power offered by modern Graphics Processing Units (GPUs) that are able to accommodate the large amounts of data required by Deep Networks. State-of-the-art image recognition networks like the ResNet family \cite{he2016resnet}, VGG networks \cite{simonyan2014vgg}, EfficientNet models \cite{tan2019efficientnet} and the recently introduced Vision Transformers \cite{dosovitskiy2020vit} require extremely large amounts of data compared to their classical Machine Learning (ML) counterparts \cite{alzubaidi2021dlreview}. 

\noindent
This characteristic requirement for large amounts of data becomes a problem in fields where data availability is low like in medical fields \cite{salehi2023cnnmedstudy}. A common approach to this problem is Transfer Learning \cite{Zhuang2021TL} which consists of pre-training a network on a large dataset like ImageNet \cite{Deng2009ImageNet} and fine tune the network on a smaller dataset that is relevant to the recognition problem at hand. In this approach, basic visual recognition features are expected to be stored in the weights of the pre-trained network which is then `transferred' to the new problem where additional layers may be used to learn the features that are specific to the custom dataset of the new problem. A shortcoming of this problem is the high computational costs associated with ImageNet pre-training and the learning of unnecessary features during the pre-training process. For example, ImageNet pre-training teaches networks about the features that distinguish a coffee mug from a specific type of fish (ImageNet classes), but when applied to a problem of X-Ray classification, these weights are not only `dead-weights' but may also be misleading and could harm the performance of the network. This calls for more robust pre-training approaches that could solve the problem of data scarcity in specific settings.

\noindent
A promising approach to overcome these problems of data hungry deep networks is to add meaningful structure to the training data of the neural networks. Bengio et al. \cite{bengio2009curriculum} were the first to formalise curriculum learning as a strategy to order training points in the order of gradually increasing difficulty. This method, instead of presenting random shuffled data-points to a network, showed more promise in terms of faster convergence and the learning of more robust features. This concept has been extended since its inception to the fields of Computer Vision and Natural Language Processing with positive results \cite{wang2021curriculumsurvey}.

\noindent
In this work, we present a combination of pre-training and curriculum learning with \textsc{Developmental PreTraining} (DPT). With the regime outlined in this paper, we hope to offer a solution for the data scarcity problem in specific fields by training networks with a regime that is designed with inspiration from early visual development in infant. The regime involves a phased pre-training approach that teaches primitive visual processing knowledge that can be transferred to various downstream classification tasks. The method is designed to not introduce irrelevant and useless features into the network's weights while also being significantly more lightweight compared to a traditional pre-training approach like with ImageNet.

\section{Related Work}
In this section, we discuss relevant related work that inspired our proposed approach of \textsc{Developmental PreTraining} as a solution to the data requirements of deep networks, especially in settings with limited availability of image data. 

\subsection{Curriculum Learning}
Curriculum Learning \cite{bengio2009curriculum} in deep neural networks draws inspiration from educational theory, guiding models through a structured progression of tasks during training. This approach aims to enhance critical metrics of a DL system like learning efficiency, and  generalization by training with handpicked, structured data-points in relation to randomly shuffled data-points in traditional training regimes. The meaningful structure of data-points is designed for the model to learn more meaningful and robust features and patterns in its data -- analogous to how a student might with a high school syllabus.

Vanilla curriculum learning \cite{bengio2009curriculum, spitkovsky2009baby} involves presenting training data in an increasing order of difficulty which is determined by a set of rules dependent on the nature of the data. Self-paced curriculum learning leverages an external set of rules to determine objective difficulty of the data-points and a metric on the model's performance on the data-points \cite{jiang2015self}.

In DPT, we hope to leverage the concept of curriculum learning to choose the nature of data in the pre-training regime in a sequentially meaningful manner with the hopes of the model learning enough primitive representations for visual processing that can be transferred to specialised tasks.

\subsection{Cognitive Neuroscience and Visual Development}\label{subsec_cogneuro_visual}
In order to develop a pre-training curriculum for \textsc{DPT}, we decided to borrow insights from early visual development. We are trying to equip our deep networks with very basic visual information processing abilities that can be transferred to a range of tasks -- similarly how an infant's inductive biases allow it to learn visual recognition tasks early on in their lives. 

Studying early visual development \cite{zaadnoordijk2022infantlessons, smith2017developmental} in humans holds much promise to the development of robust and generalisable machine learning applications in the field of vision. One common feature in early visual is a phased 'curriculum' prominent in visual learning. This curriculum starts from ingrained priors in the infant like a bias for edges \cite{linsley2020recurrentedge} to the varying set of scenes that the infant is capable of viewing determined by its physical mobility at that stage in development \cite{zaadnoordijk2022infantlessons, smith2017developmental}.

Additionally, Vogelsang et al. \cite{vogelsang2018potential} also emphasise the effects of the gradual improvement of initial low visual acuity in newborn infants on their visual development. The authors run CNN simulations to show faster convergence when trained with blurry images at the beginning and normal images towards the end. This implicit curriculum forces the visual systems to learn low-level features over larger spatial fields at the beginning and transition to more fine-grained features with reducing blur in the image. This process was found to introduce representations in the network that are more robust and generalisable to unseen images. 

This line of work introduces potential additions for a curriculum-based pre-training approach. In DPT, we prioritise a phased transition in our pre-training regime that is a common feature in the works cited in this section.

\section{Proposed DPT Regime}
In our \textsc{Developmental PreTraining} (DPT) approach, we incorporate features from early visual development in humans to create a novel curriculum-based pre-training regime. In this section, we outline, in detail, the components of this regime.

\subsection{Choice of Data}
The DPT approach, in direct contrast with the traditional approach of pre-training on a large dataset like ImageNet \cite{Deng2009ImageNet}, needs to be minimal in the visual representation it learns while also being useful. This allows the network to possess knowledge of basic visual processing that can then be utilised by downstream tasks using means like Transfer Learning. For this reason, we turned to cognitive neuroscience and early visual development in infants (see subsection \ref{subsec_cogneuro_visual}) for inspiration on the data to use in the pre-training regime.

\begin{figure}
  \centering
  \includegraphics[width=\textwidth]{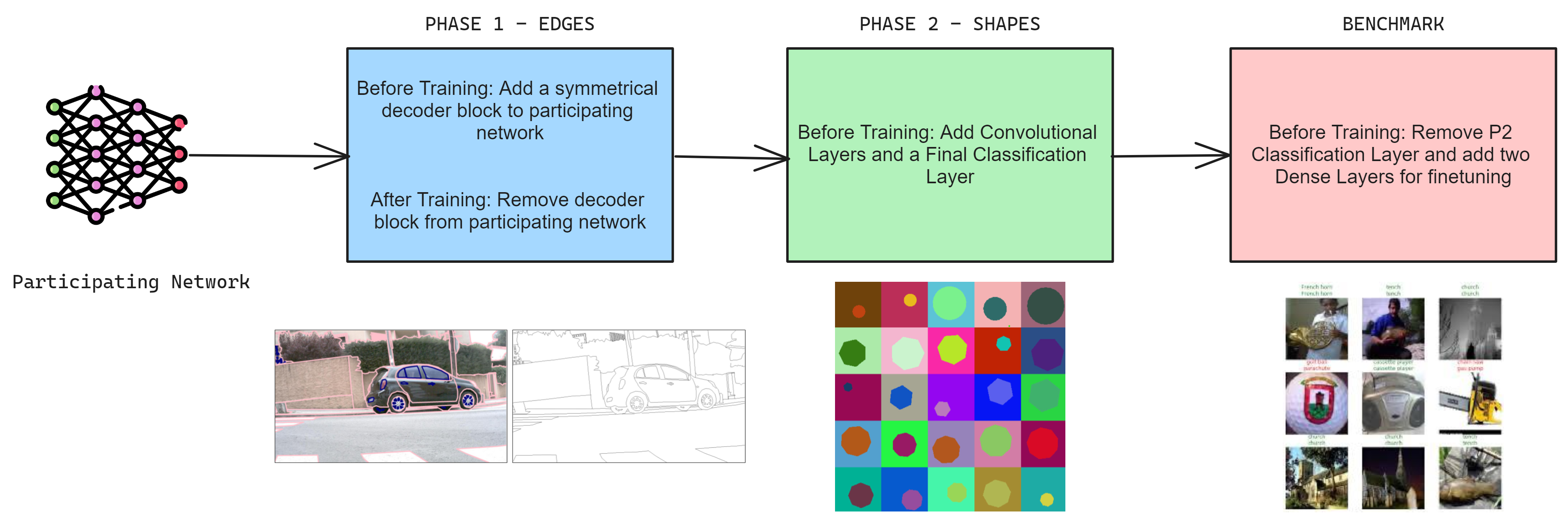}
  \caption{Overview of the DPT phases. Phase 1 teaches the participating network edge detecting representations, Phase 2 teaches the network shape recognising representations. The network is then benchmarked on a dataset containing real-world objects.}
  \label{fig:fig1}
\end{figure}

We split DPT into two phases (with plans of extending it to multiple phases in future work) where we exposed the network with a handpicked data curriculum designed to ingrain meaningful priors. Our first phase leveraged the notion that the human visual system comes ingrained with the ability of finding strong continuous lines and contiguous surfaces which are attributed to visual cortical micro-circuitry in our early brains \cite{linsley2020recurrentedge}. This inductive bias for edges in the early human visual system was the focus of \textbf{Phase 1}. DPT then transition from edges to simple structures consisting of edges - basic shapes in \textbf{Phase 2}. This allowed the network to learn representations of shapes formed by a collection of lines. The features chosen to train in DPT were designed to be primitive and low-level such that the knowledge can be transferred to any and all fields. An overview of the phases can be found in \ref{fig:fig1}.

\subsection{Phased Pre-Training}
\subsubsection*{Phase 1}
The first phase of DPT was designed to simulate the bias of early human visual systems to edges. This was achieved with the help of an edge detection task.
\paragraph{Architectural Changes}The network participating in the DPT regime would be pre-pended to a decoder block where the network would function as the encoder. This auto-encoder setup was used to perform an edge-detection task with the dataset detailed below. Once training was completed, the deconvolutional layers in the decoder block are discarded and the convolutional layers are carried forward to the second phase.
\paragraph{Dataset}Phase 1 involves training the participating network on the BIPEDv2 Dataset \cite{soria2023bipedv2}. These outdoor images are originally 1280$\times$720 (resized to 256$\times$256 for DPT) pixels each with ground truths that are annotated by experts in the computer vision field. There are 200 training images and 50 testing images.

\subsubsection*{Phase 2}\label{subsec-p2}
The second phase of DPT was designed to extend the representations learnt by the convolutional layers of the participating network to a shape recognition task. This natural, difficulty-based progression was inspired by the previously-mentioned Curriculum Learning techniques.

\paragraph{Architectural Changes}The convolutional layers with optimised weights for edge detection from Phase 1 are appended with more convolutional layers and a final classification layer. The output layer has 9 neurons denoting the 9 possible classes in the shape dataset from below. 

\paragraph{Dataset}A Geometric 2D shape dataset \cite{el2020shapes2d} was used for Phase 2. There are 9 classes in the data where each class represents a different 2D geometric shape (Triangle, Square, Pentagon, Hexagon, Heptagon, Octagon, Nonagon, Circle and Star). Each class consisted of 100,000 images with dimensions of 256$\times$256.

\subsection{Benchmarking}
Upon the completion of the pre-training phases, we benchmarked our model with a vanilla, control model of the same architecture on a real-world dataset.
\paragraph{Architectural Changes}The classification layer from Phase 2 is discarded and replaced with two new Dense layers. First is a hidden layer for fine tuning and the second is the classification layer.
\paragraph{Dataset}The Imagenette dataset \cite{imagenette} was chosen for the benchmarking. The dataset is a subset of ImageNet with 10 of the easily classified classes (tench, English springer, cassette player, chain saw, church, French horn, garbage truck, gas pump, golf ball, parachute). Imagenette was chosen to evaluate the model's capability of recognising complex features found in everyday objects.

\section{Results and Analysis}
In our internal testing of the DPT regime, we set up a CNN from scratch. Note that the pre-training regime could also be carried out for large, state-of-the-art deep networks too.

\begin{figure}
  \centering
  \includegraphics[width=0.6\textwidth]{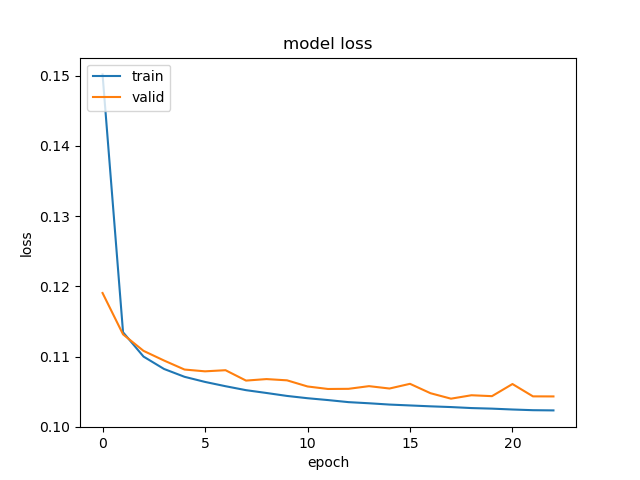}
  \caption{Phase 1 - The figure shows successful learning of edge detection by our participating network.} 
  \label{fig:fig2}
\end{figure}

For Phase 1, our network was able to quickly converge to satisfactorily low binary cross-entropy values between the predicted edge maps and the ground truth values. This can be verified in Figure \ref{fig:fig2}. In 10-15 epochs, the model was able to reach a stable loss value.

\begin{figure}
  \centering
  \includegraphics[width=0.6\textwidth]{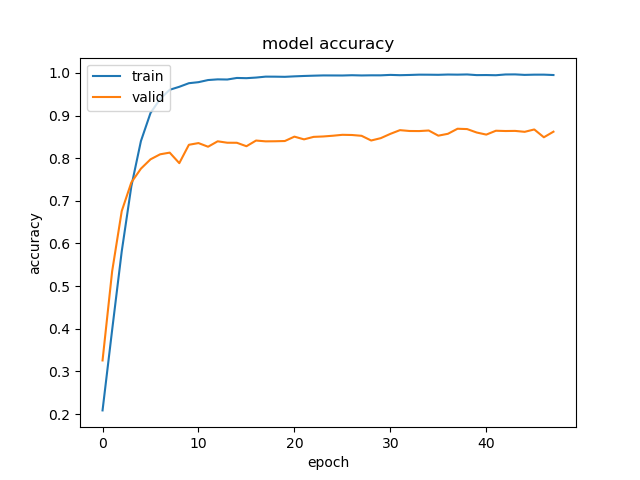}
  \caption{Phase 2 - The figure shows successful learning of shape recognition} 
  \label{fig:fig3}
\end{figure}

In Phase 2, our network was modified as per the specifications provided in Section \ref{subsec-p2}. When trained with the Shapes2D dataset, the network was able to converge to a near-perfect accuracy in about 10 epochs as seen in \ref{fig:fig3}. However, this phase struggled with some over-fitting. This was circumvented to some degree with the addition of some dropout layers \cite{srivastava2014dropout} to force the network to learn more robust and generalisable representations. These layers were removed later to maintain the pre-training requirements.

\begin{figure}
  \centering
  \includegraphics[width=0.8\textwidth]{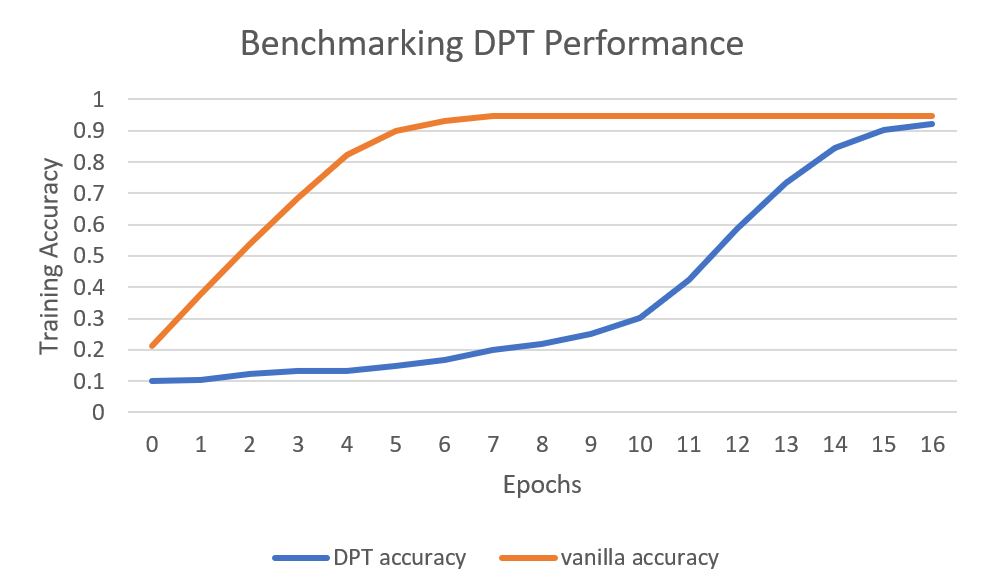}
  \caption{Benchmark - Training accuracy comparison between DPT and vanilla models} 
  \label{fig:fig4}
\end{figure}

Upon successful completion of the two phases of DPT, we proceeded to benchmark our DPT model against a Vanilla model of the same architecture but with randomly initialised weights. The benchmark was used to verify if the pre-trained weights in the network fare better for a real-world recognition task compared to a vanilla model without any special weights. \ref{fig:fig4} shows that both DPT and the vanilla models converge to similar training accuracy levels. However, these results suggest the contrary to our hypothesis that the DPT regime would lead to faster convergence compared to a non-pretrained model. 

Figure \ref{fig:fig4} suggests that the representations learned during the pre-training process weren't directly extendable to the real-world setting simulated by the Imagenette dataset. The pre-trained weights from DPT seem to be holding the model back instead of boosting performance. This could be due to the sub-optimal configuration of the phases that led to overfitting on the phases in each dataset.

\section{Conclusion}
Over the course of this study, we attempted to design a pre-training scheme to address the problem of data scarcity in certain fields. We drew inspiration from Curriculum Learning and early visual development of infants to design our \textsc{Developmental PreTrainng} regime. A phased syllabus of increasingly difficult primitive features was presented to the participating network in order to teach basic, low-level features necessary for visual processing. Although typical ImageNet pre-training approaches achieve this (and more), its large computational cost poses an impediment for those training networks on local machines. Another drawback of traditional pre-training is the learning of features that are 'un-transferable' to a novel dataset. For these reasons, we developed DPT with an emphasis on low-level features like edges and shapes to be learnt by the network using lighter datasets so that they can be used in downstream recognition problems later on.

Unfortunately, our experimentation involving the designed system did not yield expected results. Although the networks were able to converge similar performance levels, they were not able to converge at a faster rate as expected. The hypothesis was that exposure to low-level features would provide a shortcut for the DPT model during training for the benchmark on the Imagenette dataset compared to the control model with random weights. However, the pre-trained weights seemed to pose as an obstacle instead of a shortcut for training on the benchmark dataset. This could be because the DPT may have led to overfitting on the phase-specific dataset that caused the model to learn features that were not robust or generalisable.

\section{Future Work}
The shortcomings of the paper despite the related work and background for the problem suggests that a lot of work remains to be done in this area of curriculum-based pre-training before it can be deployed in the real world. We have a lot of ideas on how this work could be extended to potentially find a regime that acts as a sustainable competitor to traditional pre-training approaches.

Firstly, the optimal datasets need to be selected for a pre-training. The current regime of edge detection and shape recognition could be extended to complex object (like clothing items \cite{xiao2017fashion} or face recognition \cite{cao2018vggface2}). It is important to note that an emphasis should be placed on lightweight datasets as a primary reason to pursue this line of work is to minimize computational costs for model training while also enjoying faster convergence due to pre-learned features. Additionally, these phases should also be dealt with care so as to not cause overfitting on these phase-specific dataset. This will harm the training process of the final, downstream task. Measures like inserting dropout and batch normalisation layers could be explored and studied.

Further, pre-training approaches should lead to models that are versatile and can be applied to various fields. For this reason, DPT needs to be benchmarked across various datasets after pre-training that ranges from medical data to satellite data. Moreover, the DPT models need to be compared with the models that were pre-trained with traditional approaches. To be successful, the DPT models need to show similar rates of convergence compared to the other pre-trained models. It would also be helpful to study the actual savings in computation between DPT and traditional pre-training techniques. Finally, it would be very interesting to analyse the robustness of features learned during the DPT regime compared to traditional pre-trained models as well as models without any pre-training. Adversarial robustness could imply that the pre-training regime of DPT is likely learning visual features that are adjacent to those learnt by our visual systems.
\bibliographystyle{unsrt}  
\bibliography{references}

\end{document}